# Crowdsourcing Control: Moving Beyond Multiple Choice

**Christopher H. Lin   Mausam   Daniel S. Weld**
Department of Computer Science and Engineering
University of Washington
Seattle, WA 98195
{chrislin,mausam,weld}@cs.washington.edu

## Abstract

To ensure quality results from crowdsourced tasks, requesters often aggregate worker responses and use one of a plethora of strategies to infer the correct answer from the set of noisy responses. However, all current models assume prior knowledge of all possible outcomes of the task. While not an unreasonable assumption for tasks that can be posited as multiple-choice questions (*e.g.* $n$-ary classification), we observe that many tasks do not naturally fit this paradigm, but instead demand a free-response formulation where the outcome space is of infinite size (*e.g.* audio transcription). We model such tasks with a novel probabilistic graphical model, and design and implement LAZYSUSAN, a decision-theoretic controller that dynamically requests responses as necessary in order to infer answers to these tasks. We also design an EM algorithm to jointly learn the parameters of our model while inferring the correct answers to multiple tasks at a time. Live experiments on Amazon Mechanical Turk demonstrate the superiority of LAZYSUSAN at solving SAT Math questions, eliminating 83.2% of the error and achieving greater net utility compared to the state-of-the-art strategy, majority-voting. We also show in live experiments that our EM algorithm outperforms majority-voting on a visualization task that we design.

## 1 Introduction

Crowdsourcing marketplaces (*e.g.*, Amazon Mechanical Turk) continue to rise in popularity. Hundreds of thousands of workers produce a steady stream of output for a wide range of jobs, such as product categorization, audio-video transcription and interlingual translation. Unfortunately, these workers also come with hugely varied skill sets and motivation levels. Ensuring high quality results is a serious challenge for all requesters.

Researchers have studied quality control extensively for the case of simple binary choice (or multiple choice) questions. A common practice is to ask multiple workers and aggregate responses by a majority vote [Snow et al., 2008]. Several extensions have been proposed that track the ability of individual workers while estimating the inherent difficulty of questions [Dai et al., 2010, Whitehill et al., 2009]. These methods typically outperform majority vote and achieve a much higher accuracy.

A key drawback of prior decision-theoretic approaches to quality control is the restriction to multiple choice questions, *i.e.*, jobs where every alternative answer is known in advance and the worker has to simply select one. While many tasks can be formulated in a multiple-choice fashion (*e.g.* $n$-ary classification), there a large number of tasks with an unbounded number of possible answers. A common example is completing a database with workers' help, *e.g.*, asking questions such as "Find the mobile phone number of Acme Corporation's CEO." Since the space of possible number of answers is huge (possibly infinite), the task interface cannot explicitly enumerate them for the worker. We call these tasks *open questions*.

Unfortunately, adapting multiple-choice models for open questions is not straightforward, because of the difficulty with reasoning about unknown answers. Requesters, therefore, must resort to using a majority-vote, a significant hindrance to achieving quality results from these more general open questions.

Our paper tackles this challenging problem of modeling tasks where workers are free to give any answer. As a first step, we restrict these tasks to those which have exactly one correct answer. We make the following contributions:

- We propose a novel, probabilistic model relating the accuracy of workers and task difficulty to worker responses, which are generated from a countably infinite set.

- We design a decision-theoretic controller, LAZYSUSAN, to dynamically infer the correct answer to a task by only soliciting more worker responses when necessary. We also design an Expectation-Maximization (EM) algorithm to jointly learn the parameters of our model while inferring the correct answers to multiple tasks at a time.

- We evaluate variations of our approach first in a simulated environment and then with live experiments on Amazon Mechanical Turk. We show that LAZYSUSAN outperforms majority-voting, achieving 83.2% error reduction and greater net utility for the task of solving SAT Math questions. We also show in live experiments that our EM algorithm outperforms majority-voting on a visualization task that we design.

## 2  Background

There exist many models that tackle the problem of inferring a correct answer for a task with a finite number of possible answers. Our work is based on the model of of Dai *et al.* [Dai et al., 2010], which we now review. Their model assumes that there are exactly 2 possible answers (binary classification). Let $d \in [0, 1]$ denote the inherent *difficulty* of completing a given task, and let $\gamma_w \in [0, \infty)$ be worker $w$'s innate proneness to *error*. The accuracy of a worker on a task, $a(d, \gamma_w)$, is defined to be the probability that she produces the correct answer using the following model: $a(d, \gamma_w) = \frac{1}{2}\left(1 + (1-d)^{\gamma_w}\right)$. As $d$ and $\gamma$ increase, the probability that the worker produces the correct answer approaches 0.5, suggesting that she is guessing randomly. On the other hand, as $d$ and $\gamma$ decrease, $a$ approaches 1, when the worker will deterministically produce the correct answer.

Dai *et al.* couple this model with a utility function, which describes the worth of a correct answer, to define a Partially Observable Markov Decision Process (POMDP) that outputs a policy for TURKONTROL, their decision-theoretic controller for crowdsourced tasks. A significant limitation of this model is its inability to handle tasks with an infinite number of possible answers.

In order to address this limitation, we use the Chinese Restaurant Process [Aldous, 1985], a discrete-time stochastic process that generates an infinite number of labels ("tables"). Intuitively, the process may be thought of as modeling sociable customers who, upon entering the restaurant, decide between joining other diners at a table or starting a new table. The greater the number of customers sitting at a table, the more likely new customers will join that table.

Formally, a Chinese Restaurant $R = (T, f, \theta)$ is a set of occupied tables $T = \{t_1, \ldots, t_n | t_i \in \mathbb{N}\}$, a function $f: T \to \mathbb{N}$ that denotes the number of customers at each table $t_i$, and a parameter $\theta \in \mathbb{R}^+$. Imagine that a customer arrives at the restaurant. He can either choose to sit at one of the occupied tables, or at a new empty table. The probability that he chooses to sit at an occupied table $t \in T$ is

$$C_R(t) = \frac{f(t)}{N + \theta}$$

where $N = \sum_{t \in T} f(t)$ is the total number of customers in the restaurant. The probability that he chooses to begin a new table, or, equivalently, the probability that he chooses not to sit at an occupied table is

$$NT_R = \frac{\theta}{N + \theta}$$

Note that this probability is not equivalent to the probability that the new customer chooses to sit at a *specific* unoccupied table $t \in \mathbb{N} \setminus T$. Since there are an infinite number of unoccupied tables, the total sum of these probabilities, should they be defined this way, would be unbounded.

$\theta$ is a parameter that defines how attractive unoccupied tables are at this restaurant. As $\theta$ grows, a new customer becomes more likely to sit by himself at a new empty table.

## 3  Probabilistic Model

We seek to develop a probabilistic model of workers on tasks that have a countably infinite solution space. As a first step, we focus on tasks that have exactly one correct answer (*e.g.* the examples mentioned in the introduction). We also assume that given the correct answer and problem difficulty, the worker's capability to produce the correct answer is independent of all the previous workers' responses. However, even when conditioning on $v$ and $d$, wrong answers *are* dependent on previous workers' responses (because common mistakes are often repeated). Finally, we assume that workers are not adversarial and do not collaborate with each other.

Dai *et al.*'s model is unable to solve this problem. When one tries to extend their model to tasks with an infinite number of possible solutions, several issues arise from the difficulty of assigning an infinite number of probabilities. For instance, since their model assigns a probability to each possible solution, the naive extension of attempting to place a uniform distribution over the space of solutions is impossible.

Additionally, a good model must consider correlated errors [Grier, 2011]. For instance, consider a task that asks a worker to find the mobile phone number of a company's CEO. We can reasonably guess that the worker might Google the company name, and if one examined a histogram of worker responses, it would likely be correlated with the search results. A common error might be to return the company's main number rather than the CEO's mobile. Not all possible answers are equally likely, and a good model must address this fact.

The Chinese Restaurant Process meets our desiderata. Let tables correspond to possible incorrect solutions to the task (Chinese restaurant); a new worker (diner) is more likely to return a common solution (sit at a table with more people) than a less common solution. We now formally define our extension of Dai *et al.*'s model to the case of unbounded possible answers.

We redefine the accuracy of a worker for a given task, $a(d, \gamma_w)$, to be:

$$a(d, \gamma_w) = (1-d)^{\gamma_w}$$

As a worker's error parameter and/or the task's difficulty increases, the probability the worker produces the correct answer approaches 0. On the other hand, as the stated parameters decrease, $a$ approaches 1, meaning the worker always produces the correct answer.

In addition to the difficulty $d$ and the worker error $\gamma_w$, let $\theta \in \mathbb{R}^+$ denote the task's *bandwagon coefficient*. The parameter $\theta$ encodes the concept of the "tendency towards a common wrong answer." If $\theta$ is high, then workers who answer incorrectly will tend to provide new, unseen, incorrect answers, suggesting that the task does not have "common" wrong answers. Contrastingly, if $\theta$ is low, workers who answer incorrectly will tend toward the same incorrect answer, suggesting that the task lends itself to the same mistakes.

Figure 1 illustrates our generative model, which encodes a Bayes Net for responses made by $W$ workers on a given task. $x_i$ is a binary random variable that indicates whether or not the $i^{th}$ worker answers correctly. It is influenced by the correct answer $v$, the difficulty parameter $d$, and the error parameter $\gamma_i$. $b_i$, the answer that is provided by the $i^{th}$ worker, is determined by $x_i$ and all previous responses $b_1, \ldots, b_{i-1}$. Only the responses are observable variables.

Let $\mathbf{B_i} = \{b_1, \ldots, b_i\}$ be the multiset of answers that workers $w_1, \ldots, w_i$ provide. Let $\mathbf{A_i} = \{a_1, \ldots, a_k\}$ be the set of unique answers in $\mathbf{B_i}$. The probability that the $i+1^{th}$ worker produces the correct answer is simply the worker's accuracy for the given task:

$$P(x_i = T|d, v) = a(d, \gamma_{i+1})$$
$$P(x_i = F|d, v) = 1 - a(d, \gamma_{i+1})$$

Then, the probability that the worker's ballot is correct is defined as

$$P(b_{i+1} = v|d, v, \mathbf{B_i}) = P(x_i = T|d, v)$$

To define the probability space of wrong answers we use the Chinese Restaurant Process. Let $f(a) = |\{b \in \mathbf{B_i} | b = a\}|$, and let $R_{i,v} = (\mathbf{A_i} \setminus \{v\}, f, \theta)$ be a Chinese Restaurant Process. Then, the probability that the worker returns a previously seen incorrect answer, $y \in \mathbf{A_i} \setminus \{v\}$ is

$$P(b_{i+1} = y|d, v, \mathbf{B_i}) = P(x_i = F|d, v)C_{R_{i,v}}(y)$$

Finally, the probability that the worker returns an unseen answer is

$$P(b_{i+1} = u|d, v, \mathbf{B_i}) = P(x_i = F|d, v)NT_{R_{i,v}}$$

Here, $u$ represents whatever the worker returns as long as $u \notin \mathbf{A_i}$. More formally, $u \in U$ where $U$ is the singleton set $\{x|b_{i+1} = x \wedge b_{i+1} \notin \mathbf{A_i}\}$. We abuse notation to simplify and elucidate:

$$P(b_{i+1} \notin \mathbf{A_i}|d, v, \mathbf{B_i}) := P(b_{i+1} = u|d, v, \mathbf{B_i})$$

The model cares only about whether it has seen a worker's answer before, not what it actually turns out to be.

### 3.1 Model Discussion

We now make several subtle and important observations.

First, our model is dynamic in the following sense. As more workers provide answers, the probabilities that govern the generation of an incorrect answer change. In particular, the parameter $\theta$ becomes less and less significant as more and more workers provide answers. In other words, as $i$ goes to infinity, the probability that a new worker provides an unseen answer, $\theta/(\theta+i)$, goes to 0. As workers provide answers, the probability mass that used to dictate the generation of a new unseen answer is slowly shifted to that which determines the generation of seen answers. See Section 5.4 for a consequence of this behavior. Although we do not believe these model dynamics completely reflect the real-world accurately, we believe our model is a good first approximation with several desirable aspects. In fact, the model dynamics we just described are able to capture the intuition that as more and more answers arrive, we should expect to see fewer and fewer new answers.

Second, certain areas of parameter space cause our model to produce adversarial behavior. In other words, there are settings of $d$ and $\theta$ for a task such that the probability a worker produces a particular incorrect answer is greater than the probability a worker provides the correct answer, on average. The following theorem, proven in the supplementary materials, makes this observation.

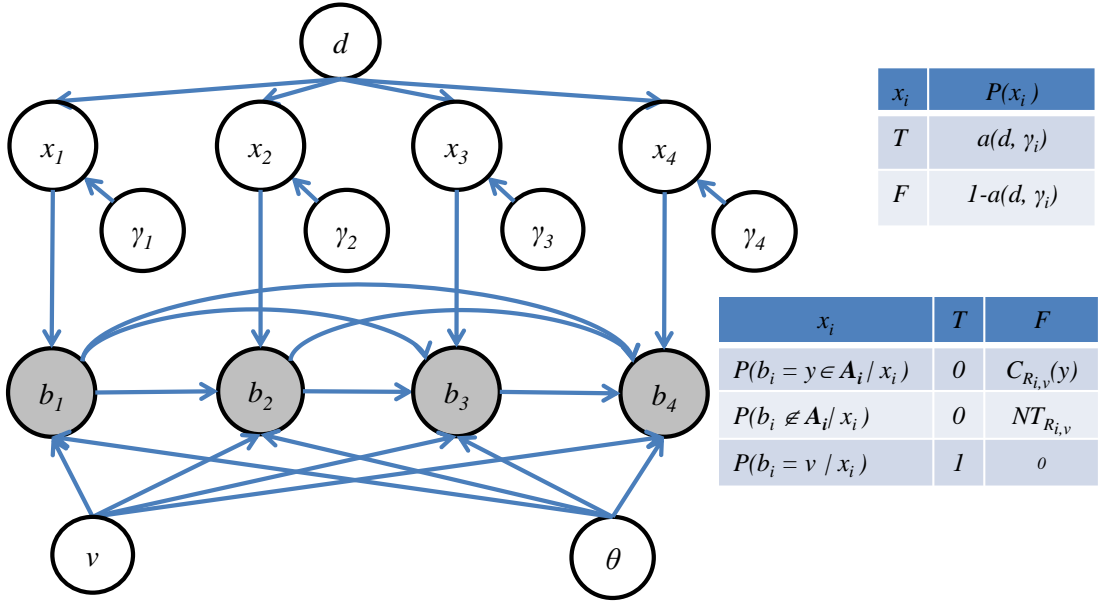

Figure 1: Whether or not worker $i$ gets it right, $x_i$, depends on his error parameter $\gamma_i$ and the difficulty of the task $d$. Worker $i$'s answer $b_i$ depends on $x_i$, the question's true answer $v$, and all the previous workers' answers $b_1, \ldots, b_{i-1}$. $R_i$ is a Chinese Restaurant Process defined by $\mathbf{B_i}$. This figure shows 4 workers. The $b_i$ are the only observed variables.

**Theorem.** *Suppose the difficulty, $d$, is fixed and all workers' $\gamma$ are equal. Then, $\theta < \frac{1-2(1-d)^\gamma}{(1-d)^\gamma}$ if and only if the expected probability the next worker returns the first-seen incorrect answer is greater than the probability the next worker returns the correct answer.*

We note that the theorem only considers the first-seen incorrect answer since the behavior of the Chinese Restaurant Process is such that the first-seen incorrect answer is generated with the highest expected probability. Thus, if the expected probability of generating the correct answer exceeds that of the first-seen incorrect answer, the model will not produce adversarial behavior.

Finally, while the purpose of this paper is to address open questions with infinite answer spaces, we note that Polya's Urn Scheme, the finite version of the Chinese Restaurant Process, applies equally well to finite answer spaces with many answer choices (multiple-choice tasks).

## 4 A Decision-Theoretic Agent

We now discuss the construction of our decision-theoretic controller, LAZYSUSAN. Our control problem is as follows. Given an open question as input, the goal is to infer the correct answer. At each time-step, an agent can choose one of two actions. It can either stop and submit the most likely answer, or it can create another job and receive another response to the task from another crowdsourced worker. The question is: How do we determine the agent's policy?

To solve this problem, first we define the *world state* of LAZYSUSAN to be the pair $(v, d)$, where $v \in \mathbb{N}$ is the correct answer of the task and $d$ is the difficulty of the task. The space of world states is infinitely large, and LAZYSUSAN cannot directly observe the world state, so it has an *agent state* $\mathbf{S}$, which at time $i$, is the set of tuples, $\mathbf{S} = \{(v, d) | v \in \mathbf{A_i} \cup \{\bot\} \wedge d \in [0, 1]\}$. Here, $\bot$ represents the case when the true answer has not been seen by the agent so far. In order to keep track of what it believes to be the correct answer $v$, it maintains a *belief*, which is a probability distribution over $\mathbf{S}$.

The agent state allows us to fold an infinite number of probabilities into one, since to compute the belief, one only needs to calculate $P(v = \bot, d | \mathbf{B_i})$, the probability that no workers have provided the correct answer yet given $\mathbf{B_i}$, instead of $P(v = u, d | \mathbf{B_i})$ for all possible unseen answers $u \in \mathbb{N} \setminus \mathbf{A_i}$.

### 4.1 Belief Update

We now describe specifically how LAZYSUSAN updates its posterior belief $P(v, d | \mathbf{B_i}; i, k)$ after it receives its $i^{th}$ ballot $b_i$. Here, $k = |\mathbf{A_i}|$ is the number of unique responses it has received. By Bayes' Rule, we have

$$P(v, d | \mathbf{B_i}; i, k) \propto P(\mathbf{B_i} | v, d; i, k) P(v, d; i, k)$$

| Symbol | Meaning |
| --- | --- |
| $\mathbf{A_i}$ | The set of unique responses made by workers $1,\ldots,i$. |
| $\mathbf{B_i}$ | The multiset of responses made by workers $1,\ldots,i$ ($\{b_1,\ldots,b_i\}$) |
| $b_i$ | Worker $i$'s response to the task |
| $C_C$ | The value of a correct answer |
| $C_{R_{i,v}}(y)$ | The probability of a worker producing incorrect answer $y$ in restaurant $R_{i,v}$. |
| $C_W$ | The value of an incorrect answer |
| $d$ | Difficulty of task |
| $\gamma_i$ | Worker $i$'s error parameter |
| $\theta$ | A task's bandwagon coefficient; Chinese Restaurant Process parameter |
| $k$ | The number of unique worker responses ($|\mathbf{A_i}|$) |
| $i$ | Number of ballots received so far |
| $Q(\mathbf{B_i}, \texttt{action})$ | The utility of taking $\texttt{action}$ with belief $\mathbf{B_i}$ |
| $NT_{R_{i,v}}$ | The probability of a worker producing an unseen answer in restaurant $R_{i,v}$. |
| $R_{i,v}$ | An instance of the Chinese Restaurant Process instantiated using $\mathbf{B_i}$ and correct answer $v$. |
| $U(\mathbf{B_i})$ | The utility of a belief derived from $\mathbf{B_i}$. |
| $V(a)$ | The value of an answer $a$ |
| $v$ | Correct answer of task |
| $x_i$ | Did worker $i$ answer the task correctly |

Table 1: Summary of notation used in this paper

The likelihood of the worker responses $P(\mathbf{B_i}|v,d;i,k)$ is easily calculated using our generative model:

$$P(\mathbf{B_i}|v,d;i,k) = \prod_{j=1}^{i} P(b_j|v,d,\mathbf{B_{j-1}})$$

The prior on the correct answer and difficulty can be further reduced:

$$P(v,d;i,k) = P(v|d;i,k)P(d;i,k)$$

The prior we must compute describes the joint probability of the correct answer and difficulty given $i$ responses and $k$ distinct responses. Notice that for all $a \in \mathbf{A_i}$, we do not know $P(v=a|d;i,k)$. However, they must be all the same, because knowing the difficulty of the task gives us no information about the correct answer. Therefore, we must only determine the probability the correct answer has yet to be seen given $d, i$, and $k$. We propose the following model:

$$P(v = \perp |d;i,k) = d^i$$

This definition is reasonable since intuitively, as the difficulty of the task increases the more likely workers have not yet provided a correct answer. On the other hand, as the number of observations increases, we become more certain that the correct answer is in $\mathbf{A_i}$.

Finally, we model $P(d;i,k)$. Consider for the moment that workers tend to produce answers that LAZYSUSAN has seen before ($\theta$ is low). Intuitively, as $k$ approaches $i$, the difficulty should grow, because the agent is seeing a lot of different answers when it shouldn't, and as $k$ approaches 1, the difficulty should become smaller, because everyone is agreeing on the same answer.

We choose to model $P(d;i,k) \sim Beta(\alpha,\beta)$ and define $\alpha \geq 1$ and $\beta \geq 1$ as

$$\alpha = \left((i-1)\frac{k}{i}+1\right)^{\frac{1}{\theta}}$$
$$\beta = \left((1-i)\frac{k}{i}+i\right)^{\theta}$$

First note that the bulk of a Beta distribution's density moves toward 1 as $\alpha$ increases relative to $\beta$, and toward 0 as $\beta$ increases relative to $\alpha$. Thus, as $\beta$ increases, difficulty decreases, and as $\alpha$ increases, difficulty increases. Consider the case when $\theta = 1$. As $k$ approaches $i$, $\alpha$ approaches $i$ and $\beta$ approaches 1, causing LAZYSUSAN to believe the difficulty is likely to be high, and as $k$ approaches 1, LAZYSUSAN believes the difficulty is likely to be low. This behavior is exactly what we desire.

Now we consider the effect of $\theta$. Fix $i$ and $k$. As $\theta$ grows, $\beta$ increases and $\alpha$ decreases. Therefore, for a fixed multiset of observations, as people become more likely to provide unseen answers, the probability that the difficulty is low becomes greater. In other words, LAZYSUSAN needs to see a greater variety of answers to believe that the problem is difficult if $\theta$ is high.

Let $\theta_1 > \theta_2$ and consider the following scenarios: Suppose $k$ is close to $i$, so LAZYSUSAN believes, before factoring in $\theta$, that the task is difficult. If $\theta = \theta_2$, LAZYSUSAN believes with more certainty that the task is difficult than if $\theta = \theta_1$. This behavior makes sense be-cause LazySusan should expect a small $k$ if $\theta$ is small. If $\theta = \theta_2$, LazySusan should see a smaller $k$ than if $\theta = \theta_1$. If it sees a $k$ that is larger than it expects, it correctly deduces that more people are getting the question wrong, and concludes the task is more difficult. Similarly, if $k$ is close to 1, and $\theta = \theta_1$, then LazySusan believes with more certainty that the task is easy than if $\theta = \theta_2$, since even though workers tend to produce more random answers, $k$ is small.

### 4.2 Utility Estimation

To determine what actions to take, LazySusan needs to estimate the utility of each action. The first step is to assign utilities to its beliefs. Since $\mathbf{B_i}$ solely determines its belief at time $i$, we denote $U(\mathbf{B_i})$ to be utility of its current belief. Next, LazySusan computes the utilities of its two possible actions. Let $Q(\mathbf{B_i}, \texttt{submit})$ denote the utility of submitting the most likely answer given its current state, and let $Q(\mathbf{B_i}, \texttt{request})$ denote the utility of requesting another worker to complete the task and then performing optimally. Then

$$
\begin{aligned}
U(\mathbf{B_i}) &= \max\{Q(\mathbf{B_i}, \texttt{submit}), \\
&\qquad\quad Q(\mathbf{B_i}, \texttt{request})\} \\
Q(\mathbf{B_i}, \texttt{submit}) &= \sum_{a \in \mathbf{A_i}} V(a) \int_d P(v=a, d|\mathbf{B_i}; i, k) \mathrm{d}d \\
Q(\mathbf{B_i}, \texttt{request}) &= c + \sum_{a \in \mathbf{A_i}} P(b_{i+1}=a|\mathbf{B_i}) U(\mathbf{B_{i+1}}) \\
&\quad + P(b_{i+1} \notin \mathbf{A_i}|\mathbf{B_i}) U(\mathbf{B_{i+1}})
\end{aligned}
$$

where $c$ is the cost of creating another job and $P(b_{i+1}|\mathbf{B_i}) =$

$$
\sum_{a \in \mathbf{A_i}} \int_d P(b_{i+1}|v=a, d, \mathbf{B_i}) P(v=a, d|\mathbf{B_i}) \mathrm{d}d
$$

LazySusan takes as inputs $C_C$, the utility of a correct answer, and $C_W$, the utility of an incorrect answer. These values are provided by the requester to manage tradeoffs between accuracy and cost. LazySusan uses its own estimate of the correct answer to calculate $V(a)$:

$$
\begin{aligned}
a* &= \operatorname*{argmax}_{a \in \mathbf{A_i}} \int_d P(v=a, d|\mathbf{B_i}; i, k) \mathrm{d}d \\
V(a) &= \begin{cases} C_C & \text{if } a = a* \\ C_W & \text{otherwise} \end{cases}
\end{aligned}
$$

### 4.3 Worker Tracking

After submitting an answer, LazySusan updates its records about all the workers who participated in the task using $a*$. We follow the approach of Dai *et al.* [Dai et al., 2010], and use the following update rules:

1) $\gamma_w \leftarrow \gamma_w - d\epsilon$ should the worker answer correctly, and 2) $\gamma_w \leftarrow \gamma_w + (1-d)\epsilon$, should the worker answer incorrectly, where $\epsilon$ is a decreasing learning rate. Any worker that LazySusan has not seen previously begins with some starting $\overline{\gamma}$.

### 4.4 Decision Making

We note that the agent's state space continues to grow without bound as new answers arrive from crowdsourced workers. This poses a challenge since existing POMDP algorithms do not handle infinite-horizon problems in dynamic state spaces where there is no a-priori bound on the number of states. Indeed, the efficient solution of such problems is an exciting problem for future research. As a first step, LazySusan selects its actions at each time step by computing an $l$-step lookahead by estimating the utility of each possible sequence of $l$ actions. If the $l^{th}$ action is to request another response, then it will cut off the computation by assuming that it submits an answer on the $l+1^{th}$ action.

In many crowdsourcing platforms, such as Mechanical Turk, we cannot preselect the workers to answer a job. However, in order to conduct a lookahead search, we need to specify future workers' parameters for our generative model. To simplify the computation, we assume that every future worker has $\gamma = \overline{\gamma}$.

### 4.5 Joint Learning and Inference

We now describe an EM algorithm that can be used as an alternative to the working-tracking scheme from above. Given a set of worker responses from a set of tasks, $\mathbf{b}$, EM jointly learns maximum-likelihood estimates of $\boldsymbol{\gamma}, \mathbf{d}$, and $\boldsymbol{\theta}$, while inferring the correct answers $\mathbf{v}$. Thus, in this approach, after LazySusan submits an answer to a task, it can recompute all model parameters before continuing with the next task.

We treat the variables $\mathbf{d}, \boldsymbol{\gamma}$, and $\boldsymbol{\theta}$ as parameters. In the E-step, we keep parameters fixed to compute the posterior probabilities of the hidden true answers: $p(v_t|\mathbf{b}, \mathbf{d}, \boldsymbol{\gamma}, \boldsymbol{\theta})$ for each task $t$. The M-step uses these probabilities to maximize the standard expected complete log-likelihood $L$ over $\mathbf{d}, \boldsymbol{\gamma}$, and $\boldsymbol{\theta}$:

$$L(\mathbf{d}, \boldsymbol{\gamma}, \boldsymbol{\theta}) = E[\ln p(\mathbf{v}, \mathbf{b}|\mathbf{d}, \boldsymbol{\gamma}, \boldsymbol{\theta})]$$

where the expectation is taken over $\mathbf{v}$ given the old values of $\boldsymbol{\gamma}, \mathbf{d}, \boldsymbol{\theta}$.

## 5 Experiments

This section addresses the following three questions: 1) How deeply should the lookahead search traverse? 2)

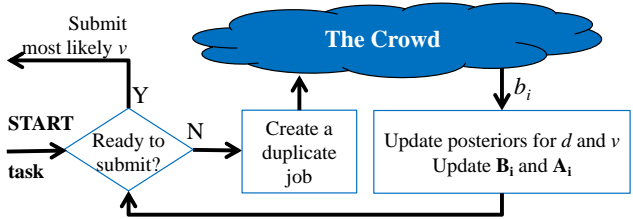

Figure 2: LAZYSUSAN's decisions when executing a task.

How robust is LAZYSUSAN on different classes of problems? 3) How well does LAZYSUSAN work in practice? and 4) How well does our EM algorithm work in practice?

To answer the first question, we compare LAZYSUSAN at different settings of lookahead depth. Then, to answer the second question, we test the robustness of LAZYSUSAN by applying it to various kinds of problems in simulation. Next, to answer the third question, we compare LAZYSUSAN to an agent that uses majority-voting with tasks that test the workers of Amazon Mechanical Turk on their SAT Math skills. Finally, to answer the fourth question, we compare our EM algorithm to a majority-voting strategy on a visualization task.

### 5.1 Implementation

Since numerical integration can be challenging, we discretize difficulty into nine equally-sized buckets with centers of $0.05, 0.15, \ldots, 0.85$, and $0.95$.

For the purposes of simulation only, each response that LAZYSUSAN receives also contains perfect information about the respective worker's $\gamma$. Thus, LAZYSUSAN can update its belief state with no noise, which is the best-case scenario.

In all cases, we set the learning rate $\epsilon = \frac{1}{m_w+1}$, where $m_w$ is the number of questions a worker $w$ has answered so far. We also set the value of a correct answer to be $C_C = 0$. Finally, we set $\bar{\gamma} = 1$, and the bandwagon coefficient $\theta = 1$ for all tasks.

### 5.2 Best Lookahead Depth

We first determine the best lookahead depth among 2, 3, and 4 with simulated experiments. We evaluate our agents using several different settings of the value of an incorrect answer: $C_W \in \{-10, -50, -100\}$. Each difficulty setting is used 10 times, for a total 90 simulations per utility setting. (9 difficulty settings × 10 = 90). We set the cost of requesting a job, $c$, from a (simulated) worker to $-1$. Our simulated environment draws worker $\gamma \in (0, 2)$ uniformly.

Figure 3 shows the results of our simulation. LAZY-

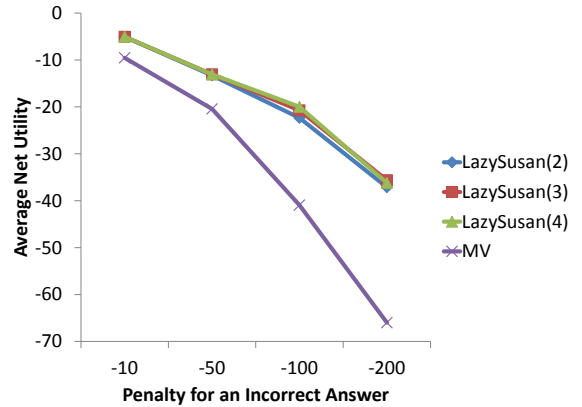

Figure 3: In simulation, our lookahead search does well at all depths.

SUSAN($l$) denotes a lookahead depth of $l$. We also examine an agent that uses a majority-of-7-vote strategy, MV. Expectedly, as the utility of an incorrect answer decreases, the average net utility achieved by each agent drops. We find that for all settings of lookahead depth, LAZYSUSAN dramatically outperforms MV. We also see that LAZYSUSAN(3) and LAZYSUSAN(4) both achieve small, but sure gains over LAZYSUSAN(2). However, LAZYSUSAN(3) and LAZYSUSAN(4) seem to do about the same. Since LAZYSUSAN(4) runs significantly slower, we decide to use LAZYSUSAN(3) in all future experiments, and refer to it as LAZYSUSAN.

### 5.3 Noisy Workers

We examine the effect of poor workers. We compare two simulation environments. In the first, we draw workers' $\gamma \in (0, 1)$ uniformly and in the second, we draw workers' $\gamma \in (0, 2)$ uniformly. Recall that $\gamma$ is the worker error parameter. Thus, the first environment has workers that are much more competent than those in the second. All other parameters remain as before. Each difficulty setting is simulated 100 times, for a total of 900 simulations per environment. The results of this experiment are shown in Table 2. When the workers are competent, LAZYSUSAN makes small gains over MV, reducing the error by 34.3%. However, when there exist noisier workers in the pool, LAZYSUSAN more decisively outperforms MV, reducing the error by 48.6%. In both cases, LAZYSUSAN spends less money than MV, netting average net utility gains of 55% and 75.9%.

### 5.4 Tasks with Correlated Errors

Next, we investigate the ability of LAZYSUSAN to deal with varying $\theta$ and $d$ in the worst case — when all the workers are equally skilled. Recall that a high $\theta$

|  | $\gamma \in (0,1)$ | | $\gamma \in (0,2)$ | |
|---|---|---|---|---|
|  | LazySusan | MV | LazySusan | MV |
| Avg Accuracy (%) | 88.7 | 82.8 | 83.3 | 67.5 |
| Avg Cost | 3.472 | 5.7 | 4.946 | 6.01 |
| Avg Net Utility | -14.772 | -22.9 | -21.646 | -38.51 |

Table 2: Comparison of average accuracies, costs, and net utilities of LazySusan and MV when workers either have $\gamma \in (0,1)$ or $\gamma \in (0,2)$

Please answer the following math question. The solution is an integer. Please enter your solution in its simplest form. (If the solution is 5, enter 5, not 5.0, and not 10/2)

What is the largest odd number that is a factor of 860?

Answer: ☐

Figure 4: An example of an SAT Math Question task posed to workers for live experiments on Mechanical Turk.

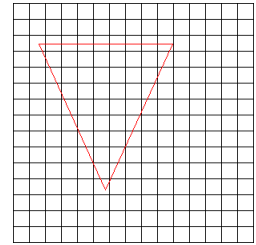

Figure 5: An example of a "triangle" task posed to workers for live experiments on Mechanical Turk. The area of this triangle, rounded down, is 38.

|  | LazySusan | MV |
|---|---|---|
| Avg Accuracy (%) | 99.25 | 95.52 |
| Avg Cost | 5.17 | 5.46 |
| Avg Net Utility | -5.92 | -9.94 |

Table 3: Comparisons of average accuracies, costs, and net utilities of LazySusan and MV when run on Mechanical Turk.

means that workers who answer incorrectly will tend to produce previously unseen answers. We consider the following variations of tasks: 1) Low difficulty, 2) High difficulty, high $\theta$, and 3) High difficulty, low $\theta$. In the first two cases, we see expected behavior. LazySusan is able to use its model to infer correct answers.

However, in the third case, we see some very interesting behavior. Since the difficulty is high, workers more often than not produce the wrong answer. Additionally, since $\theta$ is low, they also tend to produce the same wrong answer, making it look like the correct answer. If the ratio is large enough, we find that LazySusan is unable to infer the correct answer, because of the unfortunate ambiguity between high difficulty, low $\theta$ problems and low difficulty problems. In fact, as LazySusan observes more ballots, it becomes more convinced that the common wrong answer is the right answer, because of the model dynamics we mention earlier (Section 3.1). This problem only arises, however, if the model produces adversarial 0, and we see in practice that workers on Mechanical Turk generally do not exhibit such behavior.

### 5.5 Experiments on Mechanical Turk

Next, we compare LazySusan to an agent using majority-vote (MV) using real responses generated by Mechanical Turk workers. We test these agents with 134 math questions with levels of difficulty comparable to those found on the SAT Math section. Figure 4 is an example of one such task and the user interface we provided to workers. We set the utility for an incorrect answer, $C_W$, to be $-100$, because with this utility setting, LazySusan requests about 7 jobs on average for each task, and a simple binary search showed this number to be satisfactorily optimal for MV. We find that the workers on Mechanical Turk are surprisingly capable at solving math problems. As Table 3 shows, LazySusan almost completely eliminates the error made by MV. Since the two agents cost about the same, LazySusan achieves a higher net utility, which we find to be statistically significant using a Student's t-test ($p < 0.0002$).

We examine the sequence of actions LazySusan made to infer the correct answer to the task in Figure 4. In total, it requested 14 ballots, and received the following responses: 215, 43, 43, 43, 5, 215, 43, 3, 55, 43, 215, 215, 215, 215. Since MV takes the majority of 7 votes, it infers the answer incorrectly to be 43. LazySusan on the other hand, uses its knowledge of correlated answers as well as its knowledge from previous tasks that the first three workers who responded with 43 were all relatively poor workers compared to the first two workers who claimed the answer is 215. So even though a clear majority of workers preferred 43, LazySusan was not confident about the answer. While it cost twice as much as MV, the cost was a worthy sacrifice with respect to the utility setting.

Finally, we compare our EM algorithm to MV, using real responses generated by Mechanical Turk workers. We develop a "triangle" task (Figure 5) that presents workers with a triangle drawn on a grid, and asks them to find the area of the triangle, rounded down. We posted 200 of these tasks and solicited 5 responses for each. These tasks are difficult since many of the responses are off by 1. Our EM algorithm achieves an accuracy of 65.5% while MV achieves an accuracy of 54.1%.

## 6 Related Work

Modeling repeated labeling in the face of noisy workers when the label is assumed to be drawn from a known *finite* set has received significant attention. Romney *et al.* [Romney et al., 1986] are one of the first to incorporate a worker accuracy model to improve label quality. Sheng *et al.* [Sheng et al., 2008] explore when it is necessary to get another label for the purpose of machine learning. Raykar *et al.* [Raykar et al., 2010] propose a model in which the parameters for worker accuracy depend on the true answer. Whitehill *et al.* [Whitehill et al., 2009] and Dai *et al.* [Dai et al., 2010] address the concern that worker labels should not be modeled as independent of each other unless given problem difficulty. Welinder *et al.* [Welinder et al., 2010] design a multidimensional model for workers that takes into account competence, expertise, and annotator bias. Kamar *et al.* [Kamar et al., 2012] extracts features from the task at hand and use Bayesian Structure Learning to learn the worker response model. Parameswaran *et al.* [Parameswaran et al., 2010] conduct a policy search to find an optimal dynamic control policy with respect to constraints like cost or accuracy. Karger *et al.* [Karger et al., 2011] develop an algorithm based on low-rank matrix approximation to assign tasks to workers and infer correct answers, and analytically prove the optimality of their algorithm at minimizing a budget given a reliability constraint. Snow *et al.* [Snow et al., 2008] show that for labeling tasks, a small number of Mechanical Turk workers can achieve an accuracy comparable to that of an expert labeler. None of these works consider tasks that have an infinite number of possible solutions.

For more complex tasks that have an infinite number of possible answers, innovative workflows have been designed, for example, an iterative improvement workflow for creating complex artifacts [Little et al., 2009], find-fix-verify for an intelligent editor [Bernstein et al., 2010], and others for counting calories on a food plate [Noronha et al., 2011].

An AI agent makes an efficient controller for these crowdsourced workflows. Dai *et al.* [Dai et al., 2010, Dai et al., 2011] create a POMDP-based agent to control an iterative improvement workflow. Shahaf and Horvitz [Shahaf and Horvitz, 2010] develop a planning-based task allocator to assign subtasks to specific humans or computers with known abilities. We [Lin et al., 2012] create a POMDP-based agent to dynamically switch between workflows.

Weld *et al.* [Weld et al., 2011] discuss a broad vision for the use of AI techniques in crowdsourcing that includes workflow optimization, interface optimization, workflow selection and intelligent control for general crowdsourced workflows. Our work provides a more general method for intelligent control.

## 7 Conclusion & Future Work

This paper introduces LAZYSUSAN, an agent that takes a decision-theoretic approach to inferring the correct answer of a task that can have a countably infinite number of possible answers. We extend the probabilistic model of [Dai et al., 2010] using the Chinese Restaurant Process and use $l$-step lookahead to approximate the optimal number of crowdsourcing jobs to submit. We also design an EM algorithm to jointly learn the parameters of our model while inferring the correct answers to multiple tasks at a time. Live experiments on Mechanical Turk demonstrate the effectiveness of LAZYSUSAN. At comparable costs, it yields an 83.2% error reduction compared to majority vote, which is the current state-of-the-art technique for aggregating responses for tasks of this nature. Live experiments also show that that our EM algorithm outperforms majority-voting on "triangle" tasks.

In the future, we would like to address the ambiguity between high difficulty, low $\theta$ problems and low difficulty problems. We also hope to develop a generative model that does not change as responses are gathered from workers. We also hope to extend the ability of LAZYSUSAN to solving tasks that have multiple correct answers. Indeed, workers oftentimes provide the same answer in different forms (*e.g.*, in different units). Other questions may have several answers (*e.g.*, top executives often carry two mobiles). While multiple correct answers may be reduced with crisply-written instructions, an improved model may also prove useful.

**Acknowledgements**

We thank the anonymous reviewers for their helpful comments. This work was supported by the WRF / TJ Cable Professorship, Office of Naval Research grant N00014-12-1-0211, and National Science Foundation grants IIS 1016713 and IIS 1016465.